\documentclass{article}

\usepackage{latexsym}
\usepackage{amssymb}
\usepackage{amsmath}
\usepackage{amsthm}
\usepackage{booktabs}
\usepackage{enumitem}
\usepackage{graphicx}
\usepackage{color}
\usepackage{titlesec}
\usepackage[belowskip=8pt,aboveskip=2pt]{caption}
\usepackage[belowskip=0pt,aboveskip=-10pt]{subcaption}
\usepackage[title]{appendix}
\usepackage{float}
\usepackage[table,dvipsnames,svgnames,x11names]{xcolor}
\usepackage{url}

\newcommand{\BibTeX}{B\kern-.05em{\sc i\kern-.025em b}\kern-.08em\TeX}
\titlespacing*{\section}
{0pt}{2.0ex plus 1ex minus .2ex}{1.0ex plus .2ex}
\titlespacing*{\subsection}
{0pt}{2.0ex plus 1ex minus .2ex}{1.0ex plus .2ex}
\newcommand{\tba}{\textsc{AXE}}
\newcommand{\mcse}{MCSE}
\newcommand{\cdbar}{\hspace{2pt} | \hspace{2pt}}
\newcommand{\neff}{$n_{e\hspace{-0.1em}f\hspace{-0.1em}f}$}

\title{Approximate Estimation of High-dimension Execution Skill for Dynamic Agents in Continuous Domains}
\date{}
\author{
Delma Nieves-Rivera \\ din7@msstate.edu \\ Mississippi State University \\ Starkville, MS 93444 USA 
\and
Christopher Archibald \\ archibald@cs.byu.edu \\ Brigham Young University \\ Provo, UT 84602 USA
}
        
\begin{document}

\maketitle

\begin{abstract}
    In many real-world continuous action domains, human agents must decide which actions to attempt and then execute those actions to the best of their ability.
    However, humans cannot execute actions without error.    
    Human performance in these domains can potentially be improved by the use of AI to aid in decision-making. 
    One requirement for an AI to correctly reason about what actions a human agent should attempt is a correct model of that human's execution error, or skill. 
    Recent work has demonstrated successful techniques for estimating this execution error with various types of agents across different domains. 
    However, this previous work made several assumptions that limit the application of these ideas to real-world settings. 
    First, previous work assumed that the error distributions were symmetric normal, which meant that only a single parameter had to be estimated.
    In reality, agent error distributions might exhibit arbitrary shapes and should be modeled more flexibly. 
    Second, it was assumed that the execution error of the agent remained constant across all observations. 
    Especially for human agents, execution error changes over time, and this must be taken into account to obtain effective estimates. 
    To overcome both of these shortcomings, we propose a novel particle-filter-based estimator for this problem.
    After describing the details of this approximate estimator, we experimentally explore various design decisions and compare performance with previous skill estimators in a variety of settings to showcase the improvements. 
    The outcome is an estimator capable of generating more realistic, time-varying execution skill estimates of agents, which can then be used to assist agents in making better decisions and improve their overall performance.    
\end{abstract}

\section{Introduction}
In many real-world settings, agents are required to select and execute continuous actions. 
There are two factors that can influence the success of an agent in these domains. 
First, the actions that could be selected might vary in their quality.
Second, agents typically cannot execute those actions with perfect precision due to some amount of execution error that varies from agent to agent. 
As an example, imagine that you just joined a darts tournament. 
For your first game, you are assigned to play against agent A.
You get to observe a match between agent A and agent B right before your match. 
While you are observing, can you gauge the skill level of agent A? 
Specifically, can you assess the ability of A to select effective target actions?
Is it clear how accurately agent A can perform the selected actions?
Is it possible to get any insights just by observing the actions agent A is executing? 
Now picture this from a different perspective.
Imagine that you are assigned to coach agent A instead.
Can you assess agent A's abilities in terms of the questions presented above, and determine which areas should be focused on for improvement? 
Imagine having a tool to help with these types of questions.

The \emph{skill estimation problem} was introduced as a framework to address these questions \cite{extAbsAAMAS18,jair2024}. 
This prior work has focused on estimating the \emph{decision-making skill} and \emph{execution skill} of an agent, given only observations of the noisy, executed actions. 
Multiple methods have been proposed and shown to be successful at producing accurate skill estimates under different assumptions about the decision-making ability of the agent. 
In particular, the JEEDS method, which simultaneously estimates the decision-making and execution skill of an agent, has been shown to work extremely well with a wide variety of agents \cite{jair2024}. 
In this paper, we explore the question of how to translate this success to domains in which some of the assumptions made by the JEEDS are not satisfied. 
The two particular assumptions that we investigate are 1) that the execution error of an agent can be accurately modeled by a symmetric Gaussian distribution, and 2) that the execution skill of an agent is stationary and it doesn't change while the agent is being observed. 

The focus on symmetric Gaussian distributions was leveraged by the JEEDS method to allow only a single execution skill parameter to be estimated. 
However, in many cases, a higher-dimensional representation of execution skill is desired.
One case where this can occur is when actions are executed in a low dimensional space, but a more nuanced and flexible error distribution is desired. 
An example of this would be the game of darts.
The action space has two dimensions, but in real life, errors are not necessarily distributed symmetrically in the 2D action space. 
To model execution error as a multivariate Gaussian distribution with an arbitrary covariance matrix, there are now 3 parameters to estimate (the variance in the 2 principal dimensions and the correlation between them). 
This gives us now a 3-dimensional space over which to maintain a distribution and perform inference to estimate the execution skill of the observed agent. 
The JEEDS method maintained beliefs over a discretization of the single execution skill parameter, but the processing time was linear in the number of hypotheses utilized. 
Increasing to three dimensions drastically reduces the coverage of the hypotheses and/or increases the running time of the algorithm. 

Another case where a higher-dimensional representation of execution skill is desired is when the action space itself is high-dimensional.  
An example is computational billiards, where the action space has 5 dimensions \cite{arch:IJCAI09,archibald2010success}. 
As each dimension has a different scale and physical meaning, it is anticipated that execution error will be different in each dimension, resulting in 5 separate parameters to maintain a distribution over and estimate. 
In either case, the result is a need for the ability to estimate execution skill in a higher dimension. 

The focus on stationary execution skill is perhaps not an oversimplification when observations of an agent acting are gathered over the course of a limited time frame, but even then, it might be desirable to be able to see the effects of fatigue on execution skill.  
Certainly, over longer time horizons, for example to monitor the execution skill of an athlete over an entire season, or a career, some adaptation of previous methods must be made to enable them to produce time-varying skill estimates that can shed insight into an agent's changing abilities.

To perform an accurate estimation of agent skill and overcome these two limitations of previous work, in this paper, we propose a novel particle-filter-based estimator.
This new estimator provides the ability to estimate execution skill in higher dimensions, giving the designer more flexibility over how the algorithm scales.
It also provides a framework in which time-varying execution skill can be estimated. 
A final contribution of this paper is the application of this novel method to real-world pitching data from Major League Baseball (MLB), demonstrating the ability to give insight into a pitcher's ability, which is an important necessity for any AI system that might provide target action suggestions.

The remainder of the paper will proceed as follows: first, in Section \ref{sec:background}, the necessary background, notation, definitions, and related work will be provided that enable us to present the proposed method and give context to its contributions. 
Section \ref{sec:mcse} will describe the proposed skill estimation method. 
Section \ref{sec:exp-setup} will describe the experimental procedure we will use, and the results of the experiments will be presented in Section \ref{sec:exp-results}.
In Section \ref{sec:baseball}, we apply the proposed method to data from MLB pitchers and compare the results to those from a previous method.
Finally, in Section \ref{sec:conclusion} we conclude.

\section{Background} \label{sec:background}
    This section presents the relevant terminology and notation required to present the proposed method.

    \subsection{Problem Statement}
\begin{figure}
        \centering
        \includegraphics[width=0.95\columnwidth]{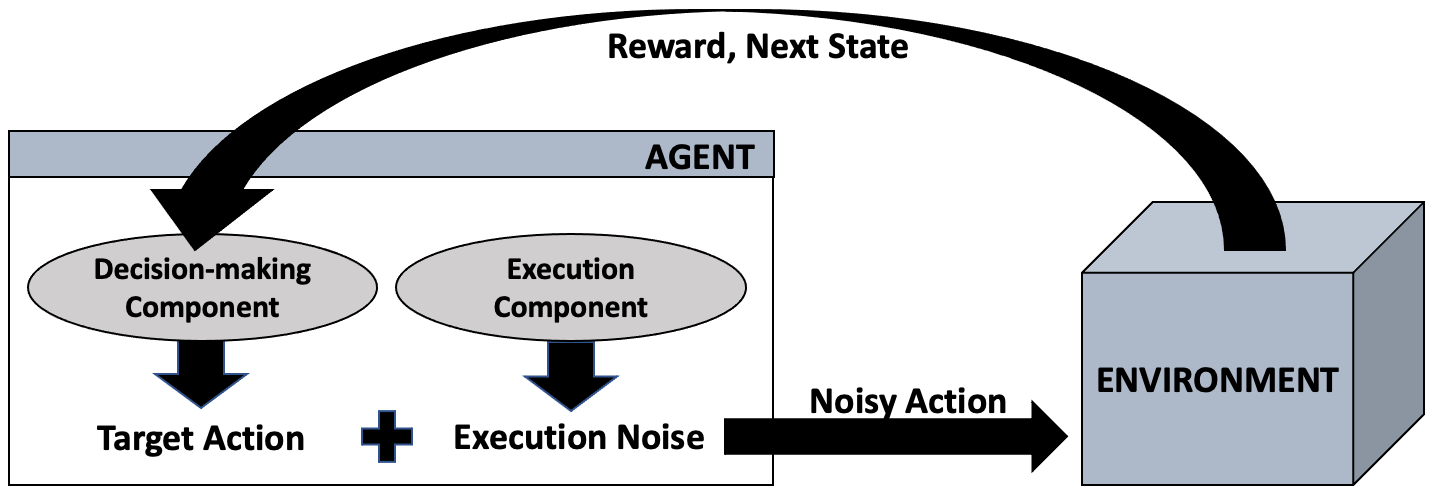}
        \caption{Agent Environment Interaction (from \cite{jair2024})}
        \label{fig:setting}
    \end{figure}

    The environments of interest for the skill estimation problem are those that feature continuous action spaces and can be modeled as Markov Decision Processes (MDPs).
    These MDPs consist of a set of states $S$, a continuous set of actions $A \subseteq \mathbb{R}^n$, a reward function $R: S \times A \mapsto \mathbb{R}$ and a transition function $P$ which specifies a distribution over potential subsequent states for any given state and action combination.
    An agent is an entity that acts in an MDP and consists of two components: one for decision-making and one for action execution. 
    The decision-making component utilizes information about the current state of the MDP and the agent's execution component to determine a target action, or more generally, a probability distribution over target actions, given the state. 
    The distribution can be of any form so long as it produces a final \textit{intended} or \textit{target} action for a given state.
    The execution component refers to how accurately an agent can execute intended actions.
    This is represented by a probability distribution over random perturbations, $\chi$, from which a perturbation is sampled and added to the target action every time the agent acts. 
    The resulting action, which is then actually executed in the MDP, is referred to as the \emph{executed} action. 
    Figure \ref{fig:setting} depicts the relationship between these components as a part of the agent's interaction with the environment.

    The concepts of \emph{decision-making skill} and \emph{execution skill} were introduced in \cite{archibald2010success} as terms to describe the quality of an agent's reasoning and the agent's execution accuracy, respectively, in our setting.
    As described above, this notion of skill gave rise to the \emph{skill estimation problem} \cite{extAbsAAMAS18,jair2024}.
    Precisely, this problem is to estimate parameters describing the agent's decision-making and execution components, given observations of the agent acting in the MDP.
    These observations contain descriptions of the state as well as the executed action, but crucially, they do \emph{not} include the target action.
    Methods have been proposed for estimating both execution skill and decision-making skill, and have been shown to be effective at producing skill estimates, under certain assumptions.
    
    \subsection{Previous Skill Estimation Methods}
    
        Several estimation methods have been proposed in the literature and shown to be effective at producing execution skill estimates.
        These methods differ in the type of information they utilize.
        The \textit{Observed Reward} (OR) method \cite{extAbsAAMAS18} focused on estimating an agent's execution skill by analyzing the rewards it receives during interactions with the environment.
        This method compares the average observed reward obtained by the agent with the rewards that would be expected from perfectly rational agents with a variety of execution skill levels. 
        The final estimate is produced by interpolation. 
        The OR method assumes that agents are perfectly rational, which limits its applicability.
        It was shown to produce accurate estimates for rational agents, but fails when agents do not make decisions optimally.
        
        To relax the perfect-rationality assumption, an alternative method, The Bayesian Approach (TBA) (later referred to as \tba{} in \cite{jair2024}), was proposed \cite{archibaldAAAI19XSkill}.
        This method models the skill estimation problem as a Bayesian network.
        Given the network, probabilistic inference can be performed to produce execution skill estimates, under the assumption that the agent would select one of a set of \emph{focal actions}, or another action uniformly at random.  
        This led to improved performance with a wider variety of agents, but required the definition of a set of focal points for a domain. 
        
        To avoid the need for any assumptions about an agent's decision-making skill, later work proposed an alternate Bayesian network that explicitly incorporates the decision-making component as a random variable \cite{jair2024}.
        The resulting \emph{Joint Estimation of Execution and Decision-making Skill} (JEEDS) method estimates both an agent's execution and decision-making skill levels simultaneously. 
        Reward information is utilized to reason about the agents' decision-making, and the executed actions give insight into execution skill.
        JEEDS was shown to produce far more accurate skill estimates for agents across the entire spectrum of rationality levels. 
        
        Our proposed method, described in Section \ref{sec:mcse}, utilizes a similar Bayesian approach as the prior methods, but differs in its modeling of the agent's skill as time-varying random variables in the dynamic Bayesian network. 
        Our proposed method also utilizes a different Bayesian inference algorithm to determine its skill estimates given the observations.
        The performance of the proposed method will be experimentally compared to JEEDS, since JEEDS is the best-performing existing method.

\subsection{Related Work} \label{sec:related}
    The topic of skill estimation has connections to many other lines of research. 
    Methods for estimating the execution error of darts players have been proposed \cite{tibshirani2011statistician,miller21darts}, but in this work, the assumption is that the target action of the player is known. 
    Recent work has proposed a novel handicap system for darts that utilizes execution skill estimates of the players \cite{chan2024throwing}.
    A previous execution skill estimation technique \cite{archibaldAAAI19XSkill} was used as a vital component in an application to baseball pitching \cite{melville2023baseball}. 
    In that paper, the interaction between pitcher and batter was modeled as a game, and equilibrium strategies for the pitcher were computed under various assumptions. 
    This gives a good example of how execution skill estimates can be leveraged within larger frameworks to provide AI assistance to humans in real-world domains. 
    Another paper focused on correlating the shape of a pitcher's error distribution to their pitching mechanics, which was done in a controlled setting where targets could be specified for the pitcher \cite{pitchingForm2017}.

    The notion of decision-making skill has been defined for players in sports \cite{araujo2009development}, and also relates to work on bounded rationality \cite{simon1972theories}.
    The specific softmax model that we utilize in this paper has been used in many other works, including the definition of quantal response equilibrium \cite{McKelvey95}. 
    It has been used to adjust the strength of AI agents for Go \cite{wu2019strength}, and to model human behavior \cite{yang2011improved}. 
    Other work has proposed methods for determining this parameter for agents, given observations of their actions in two-player zero-sum games \cite{ling2019large}.

    Estimating the properties of other agents in games is often called opponent modeling, and skill estimation can be seen as a method for determining important properties of observed agents \cite{nashed2022survey}. 
    Opponent modeling has been explored in the context of poker \cite{13aamas-implicitmodelling,15aamas-clustering,billings1998opponent}, real-time strategy games \cite{schadd2007opponent}, and $n$-player games \cite{06aaai-probmaxn}. 
    
    One could model the execution error as part of the transition function of the underlying MDP, resulting in a different MDP for each agent in the same environment. 
    One potential way to view skill estimation work is as an attempt to separate transition information that is agent-specific from environment-specific information. 
    This could potentially be helpful for transfer learning, an important problem in reinforcement learning \cite{taylor2009transfer,zhuang2020comprehensive}. 
    For example, how can a strategy developed for an agent with one execution skill level be utilized by an agent with a different execution skill level?

    Finally, our proposed skill estimation method utilizes standard Bayesian reasoning techniques \cite{AIBookRussell}. 
    Particle filters have been used extensively in robotics to track the state of a robotic system \cite{thrun2005probabilistic}. 
    One previous study used a particle filter for opponent modeling in Kuhn poker \cite{07aaai-om}. 
    The task was to infer the parameters representing the strategy being used by the opponent in the game. 
    At a high level, our approach is similar to theirs, but all the specific details: the components of the particle filter, the nature of the observations, and the whole setting, are different. 

\section{Monte Carlo Skill Estimation} \label{sec:mcse}
This section presents the details of the proposed method: \emph{Monte Carlo Skill Estimation} (\mcse{}).
\mcse{} operates within a Bayesian framework, representing agent skill levels, target actions, and executed actions as random variables. 
A probability distribution over the space of possible skill levels will be updated each time an executed action is observed. 
What differentiates \mcse{} from prior work is the fact that the skill levels are represented as \emph{time-varying} random variables in the dynamic Bayesian network, and that a particle filter is used as the inference method to reason about the involved random variables.
The remainder of this section will describe the \mcse{} method in detail. 

\subsection{Bayesian Framework}
\begin{figure}[ht]
    \centering
    \includegraphics[width=0.5\columnwidth]{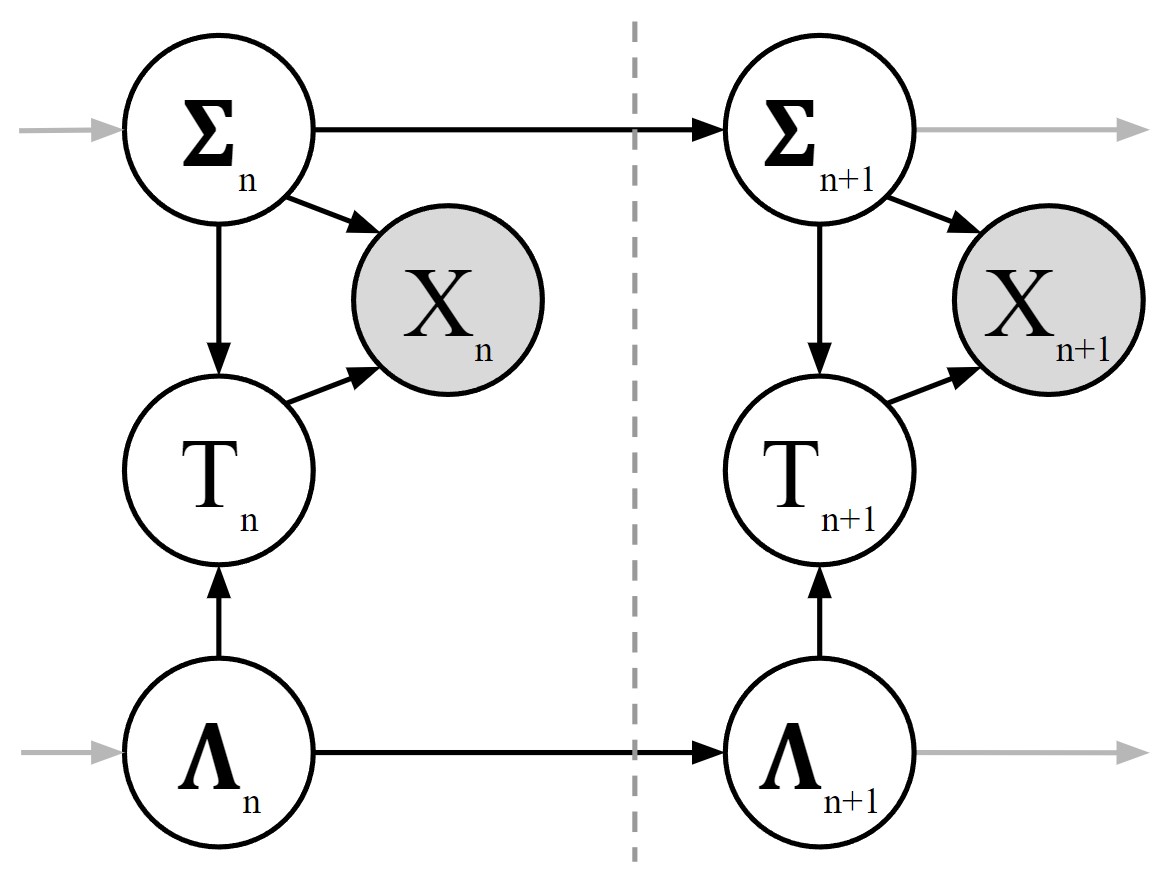}
    \caption{Dynamic Bayesian Network for \mcse{} }
    \label{fig:mcse-net}
\end{figure}

The dynamic Bayesian network used to guide inference for \mcse{} is shown in Figure \ref{fig:mcse-net}. 
It consists of the following random variables: 
    \begin{itemize}
        \item $\Sigma$ is a multivariate random variable consisting of $k$ dimensions which correspond to the execution skill parameters of the agent that are to be estimated. 
        $\Sigma$ is unobserved and isn't directly influenced by anything else.  
        Execution skill estimation methods will perform inference to update a probability distribution over possible execution skill levels, given the observations.
        $\sigma$ will be used to indicate a specific set of execution skill distribution parameters. 
        \item $\Lambda$ is a random variable corresponding to the decision-making skill of the agent. 
        This is also unobserved and isn't influenced by anything else, as it is considered an inherent characteristic of an agent.
        Decision-making skill estimation methods will infer a distribution over possible decision-making skill levels, given observations.
        $\lambda$ will be used to refer to a specific decision-making skill level.
        \item $T$ is a random variable indicating the target action for a given observation.  
        This random variable is also not directly observed, but can be influenced by the execution and decision-making skill random variables $\Sigma$ and $\Lambda$. 
        A target action for a given observation will be referred to as $t$.
        The influence of the current state on the target action will be left implicit.
        \item $X$ is a random variable indicating the action that is actually executed and observed. 
        A specific executed action will be referred to as $x$.
        This random variable is directly influenced by the target action random variable $T$, as well as the execution skill multivariate random variable $\Sigma$.
        $P(x \cdbar t, \sigma)$ indicates the true conditional distribution over executed actions given an intended action $t$ and an execution skill level $\sigma$, and will directly correspond to the execution noise distribution $\chi^{\sigma}$, centered on $t$. 
    \end{itemize}
All of the variables are indexed by the observation number, as in this work, we model agents' skill as time-varying, and the target and executed actions will in general also differ from observation to observation.
Moreover, we focus on episodic environments but we hypothesize that \mcse{} can be successfully extended to sequential domains in the same way JEEDS was extended \cite{jair2024}, by replacing the reward function with an optimal action-value function.

\subsection{Particle Filter Approach Overview}
The proposed \mcse{} method will utilize a particle filter to perform probabilistic inference over the space of possible skill levels. 
A set of particles will be maintained, and this set will represent the current probability distribution over the agent's skill.
Each particle will correspond to a complete specification of an agent's skill, including specific execution skill parameters $\sigma$ and rationality parameter $\lambda$.
Every time an executed action and state are observed, each particle will compute a weight. 
This weight corresponds to the probability of an agent with that particle's skill level executing the observed action. 
The particle will multiply its previous weight by this new weight.
Periodically, a new set of particles will be resampled from the old set, with replacement. 
When resampling occurs, the weight for each newly resampled particle will be initialized to 1. 
When a skill estimate is required, it will be computed as the weighted average value of all particle skill parameters. 
Details for each of these components will be given in the following sections.

\subsection{Execution Skill Model}
The \mcse{} method requires the ability to compute the probability of executing any specific action, given the parameters for an execution noise distribution and a target action. 
This will be represented abstractly by a probability density function (pdf) $f_{\sigma}$, where $\sigma$ represents a specific set of execution skill parameters.  
Thus, the conditional probability distribution $P(x \cdbar t, \sigma) = f_{\sigma}(x \cdbar t)$.
This pdf can be made concrete given the specific parameterized execution skill distribution that will be used in a setting. 

\subsection{Decision-making Skill Model}
The \mcse{} method uses the softmax function, shown in Equation \ref{eq:logit}, to model imperfect decision-making as a function of the rationality parameter $\lambda_b$, as was done in \cite{jair2024}. 
This function provides a distribution over target actions, given a specific state, execution skill parameters $\sigma_b$, and decision-making parameter $\lambda_b$.
Higher $\lambda_b$ values correspond to increased decision-making skill in this model.  
    
\begin{equation}\label{eq:logit}
    P(t \cdbar \sigma, \lambda) = \frac{e^{\lambda V(t,\sigma)}}{\sum_{t' \in A}e^{\lambda V(t',\sigma)}}
\end{equation}
The function $V: A \times \Sigma \mapsto \mathbb{R}$ is used to represent the expected reward for an action is a state $s$, given a target action and execution skill level.  
This is computed as $V(a, \sigma_b) = \mathbb{E}_{\varepsilon \sim \chi^{\sigma_b}}[R(s, a + \varepsilon)]$.

\subsection{\mcse{} Method Description}
The specific steps of the \mcse{} method are now provided. 
The details for some of the steps will be given in subsequent sections.

\begin{enumerate}
    \item Initialize the set of particles $B = \{b_0, b_1, \ldots, b_M\}$, where each $b_i = ( \sigma_{b_i}, \lambda_{b_i})$, with all skill parameters sampled from uniform distributions over their respective ranges. 
    \item Repeat for each observed executed action $x_n$:  
    \begin{enumerate}
        \item For each particle $b \in B$ update weight as \\ $w_b \mathrel{*}= P(x_n \cdbar \sigma_b, \lambda_b)$ using Equation \ref{eq:weight-update}
        \item Possibly resample new set of particles (see Section \ref{sec:resample})
        \item For each new particle $b \in B$ perturb particle parameters as described in Section \ref{sec:perturbation} to account for dynamic skill
    \end{enumerate}
    
\end{enumerate}

\subsubsection{Weight Computation}
The weight for a particle $b$ corresponds to $P(x_n \cdbar \sigma_b, \lambda_b)$, which is the probability of observing executed action $x_n$, given skill parameters $\sigma_b$ and $\lambda_b$. 
The formula for this weight is shown in Equation \ref{eq:weight-update}, which was derived using the structure of the dynamic Bayesian network shown in Figure \ref{fig:mcse-net} and the \mcse{} components discussed earlier. 
The derivation follows that given in \cite{jair2024}.

\begin{equation} \label{eq:weight-update}
    P(x_n \cdbar \sigma_b, \lambda_b) = \frac{1}{\sum_{t' \in A} e^{\lambda_b V(t', \sigma_b})} \sum_{t_n \in A} e^{\lambda_b V(t_n, \sigma_b)} f_{\sigma_b}(x_n \cdbar t_n)
\end{equation}

\subsubsection{Resampling Procedure} \label{sec:resample}
When the particles are resampled, a certain percentage ($r$) are resampled from the previous set, with replacement. 
Each particle is selected in proportion to its weight. 
Randomly initialized particles are then added to the new set of particles until there are again $M$ particles. 
In this paper, we investigate two methods for determining when to resample the set of particles. 
The first is the simplest, the set of particles is resampled after every observation. 
The second focuses on computing the \emph{number of effective particles} \neff{} and only resamples when \neff{} is below a certain threshold. 
\neff{} is computed by summing all the particle weights. 

\subsubsection{Particle Perturbation} \label{sec:perturbation}
When a single particle is resampled more than once, multiple copies of it are included in the set of particles. 
Without further steps, each of these particles would simply have the same weight and could not help cover the entire space of possible skill levels that the agent could have. 
In this step, the parameters of each particle are perturbed.  
This is done by independently sampling a zero-mean Gaussian distribution for each parameter. 
The standard deviation of each of these distributions is set to be a specific fraction $w_{\%}$ of the width of the corresponding parameter's range of possible values. 
We will explore the performance of \mcse{} with different values for $w_{\%}$ in Section \ref{sec:exp-results}.

\subsubsection{Generating a Skill Estimate} \label{sec:generating}
Finally, given a set of particles along with their corresponding weights, how is a skill estimate produced? 
In \mcse{} the weighted average of all the particle parameters is computed, and the skill levels specified by those average skill parameters are the estimates produced by the model. 
If an estimate is required immediately following a resampling step, then the newly random particles are not included in this weighted average. 

\section{Experimental Setup} \label{sec:exp-setup}
In this section, the experiments that were used to evaluate the \mcse{} method are described. 
These experiments were done in a simulated variant of darts, a commonly used domain for execution skill estimation. 
This simulated experimental environment allows for the accuracy of different skill estimation techniques to be evaluated, since the true skill levels of the agents are known. 

    \subsection{Experimental Domain: Two-Dimensional Darts}
    The traditional game of darts involves players throwing pointy projectiles (the darts) at a circular board. 
    The player gets points depending on which region of the board the dart sticks in. 
    The traditional game of darts has been the basis for much previous work in estimating agent skill \cite{tibshirani2011statistician,miller21darts,jair2024}.
    We utilize the \textit{two-dimensional darts} (2D-Darts) variant of the game introduced in \cite{jair2024}, where the base rewards of the traditional dartboard are randomly shuffled.
    This variation guarantees that each dartboard presented to an agent is different and challenging, enabling the agent to showcase its decision-making and execution abilities more explicitly.

    We parameterize execution skill in 2D-Darts using three parameters, which together define the covariance matrix of a bivariate Gaussian distribution. 
    The three parameters we use are $\sigma_x$, $\sigma_y$, and $\rho$, which result in a covariance matrix of 
    $\Sigma = \begin{pmatrix} 
\sigma_x^2 & \sigma_x \sigma_y \rho\\
\sigma_x \sigma_y \rho & \sigma_y^2
\end{pmatrix}
$.
We will represent by $f_{\Sigma}$ the probability density function corresponding to a bivariate Gaussian distribution with covariance $\Sigma$. 
        
    \subsection{Agents} 
    Different types of agents, each with an unique decision-making component, were utilized for the experiments.
    These are the same types used in the experiments conducted in \cite{jair2024}.
    The \textit{Rational} agent selects an optimal action that will maximize its expected reward, with respect to its execution skill level.
    The \textit{Flip} agent employs an $\epsilon$-greedy strategy, allowing it to select an optimal action w.p. $\lambda_f \in [0,1]$ or a uniform random one.   
    The \emph{Softmax} agent uses Equation \ref{eq:logit} to probabilistically select an action w. p. $\lambda_s \in [0,\infty]$.
    From the possible actions that will yield an expected reward of $\lambda_d \in [0,1]$ times the maximum possible expected, the \emph{Deceptive} agent selects the action that is the farthest from an optimal action.
    Note that the agents with an imperfect decision-making component have a single parameter ($\lambda$) to represent their level of rationality.
    For each, the higher the $\lambda$, the more rational the agent is.
    We assume that each agent has a correct model of its execution noise.

    Each of these types of agents were combined with a wide variety of execution skill levels in the experiments. 
    For the experiments conducted in \cite{jair2024}, only agents with non-changing symmetrical execution skill distributions were tested.
    In contrast, for the current work, agents with both symmetric and asymmetric distributions were used.
    In addition, experiments were conducted both when agents have stationary execution skill distributions and when they have time-varying distributions. 
    The results and findings for these experiments are presented in Section \ref{exps:multi-dimensions} and Section \ref{exps:dynamic-xskill}, respectively.
    
    \subsection{Experimental Procedure}
    The process used to conduct each experiment was as follows. 
    First, each agent being used in an experiment was initialized with the same execution skill level and a random decision-making skill level sampled from their respective range of rationality parameters. 
    Each agent then repeatedly saw a state, selected a target action, and had an execution noise perturbation drawn from its execution noise distribution added to its target action. 
    The final executed action was then observed by each of the estimation methods, which produced an estimate after each observation.
    All agents in an experiment experienced the same sequence of environment states and the same sequence of execution noise perturbations, to reduce variance in the results. 
    
    A resolution of 5.0 mm was used to discretize the action space.
    The reward function for each state was convolved with the agent’s execution noise distribution (using the same resolution) to compute the expected reward for all actions.
    Bivariate zero-mean Gaussian distributions were used for the true execution noise distributions, where $\sigma_x$ and $\sigma_y$ were selected from the range $[3.0, 150.5]$ (mm), and $\rho$ from the range $[-0.75, 0.75]$.
    The agent rationality parameter ranges were: $\lambda_f, \lambda_d \in [0.0, 1.0]$, and $\lambda_s \in [0.001, 32.0]$.

    \subsection{Evaluating Execution Skill}
    Previous estimators output a single number representing the standard deviation of the agent's symmetric execution noise distribution. 
    Estimators were compared using the mean squared error (MSE) of this estimate compared to the true standard deviation.
    The \mcse{} method produces estimates for multiple execution skill parameters so the MSE won't be as meaningful for comparison, as an error for each parameter would need to be included. 
    Instead, we use Jeffreys divergence (JD) to measure the difference between the estimated execution noise distribution and the agent's true execution noise distribution. 
    The Jeffreys Divergence between two distributions $P$ and $Q$ is defined as follows: $JD(P || Q) = D_{KL}(P || Q) + D_{KL}(Q || P)$, where $D_{KL}$ is the Kullback-Leibler (KL) divergence. 
    The KL divergence between two bivariate normal distributions can be computed in closed form from the corresponding means and covariance matrices. 
    
    \subsection{Preliminary Experiments}
    The \mcse{} method has multiple parameters that must be specified.
    These include the number of particles to use ($M$), the amount of motion model noise to add to the parameters $w_{\%}$, the resampling strategy (always or \neff{}, the percentage of particles to resample ($r$), and the \neff{} threshold to use when it is the resampling strategy, 
    A preliminary set of experiments was conducted to explore different parameter values to select good parameters for the remaining experiments. 
    These experiments only used stationary Rational agents, running for 100 observations in 2D-Darts. 
    The results of this exploration are presented in the Appendix \ref{apx:prelim}, but the general observations were that higher $M$ improved performance, $w_{\%}$ didn't appear to have much of an impact, and higher $r$ values generally led to better performance. 
 
    Based on this exploration, in the the remaining 2D-Darts experiments the \mcse{} parameter combination used was: $M=1000$, $r=0.9$, $w_{\%}=0.005$, with resampling based on \neff{} with a threshold of 0.5.
    This combination had the smallest average JD after 100 observations in the preliminary experiments. 
    We will refer to this estimator as \mcse{}.

    The selected \mcse{} method was compared to the JEEDS estimator from previous work.
    In the 2D-Darts experiments, JEEDS utilized 33 hypothesis skill levels for both decision-making and execution skill, resulting in 1,089 combined hypotheses. 
    This number was chosen so that JEEDS would have approximately the same number of hypothesis samples as \mcse{}.
    The decision-making skill level hypotheses for JEEDS were logarithmically distributed.
    All JEED beliefs were initialized to be uniform over the space of possible skill parameters.

   \begin{figure}[ht]
        \centering
        \begin{subfigure}[b]{0.3\columnwidth}
        \centering
            \includegraphics[width=\textwidth]{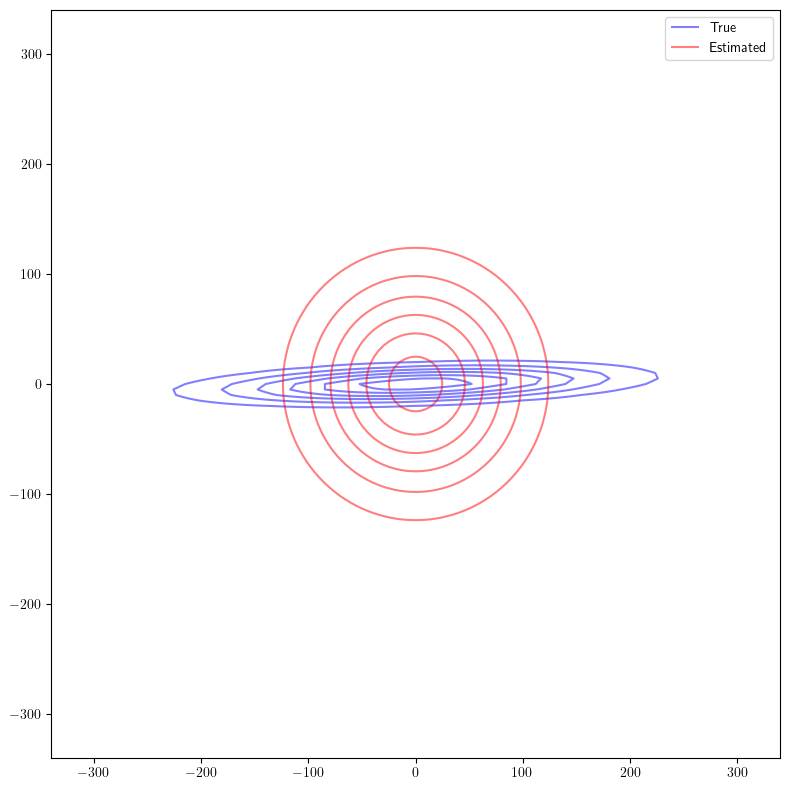}
            \label{fig:d1}
             \caption{$JD = 19.5637$}
        \end{subfigure}
        \hspace{2em}
        \begin{subfigure}[b]{0.3\columnwidth}
        \centering
            \includegraphics[width=\textwidth]{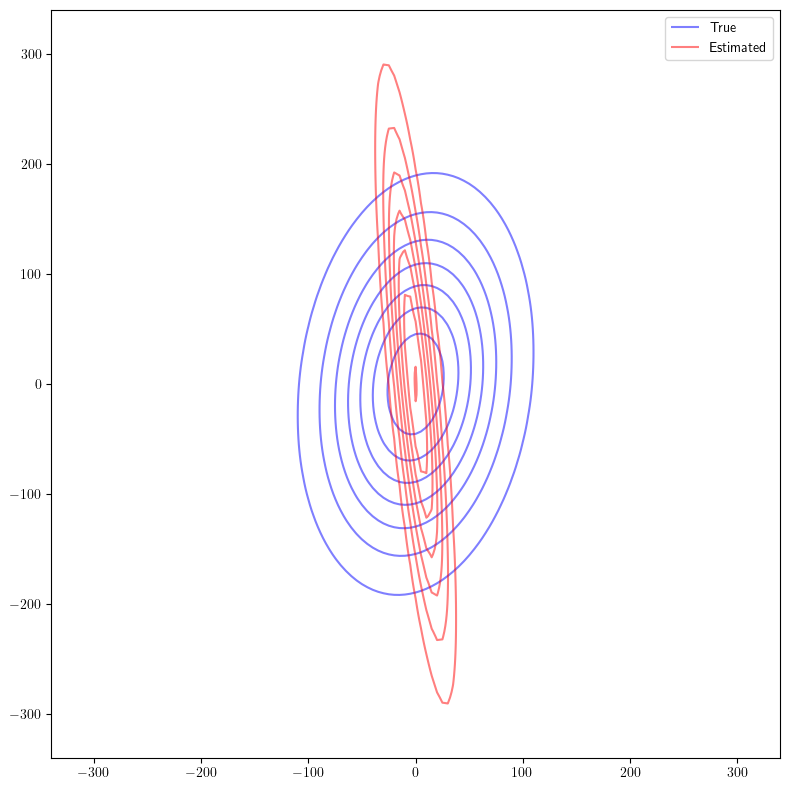}
            \label{fig:d2}
            \caption{$JD = 9.3760$}
        \end{subfigure}
        
        \begin{subfigure}[b]{0.3\columnwidth}
        \centering
            \includegraphics[width=\textwidth]{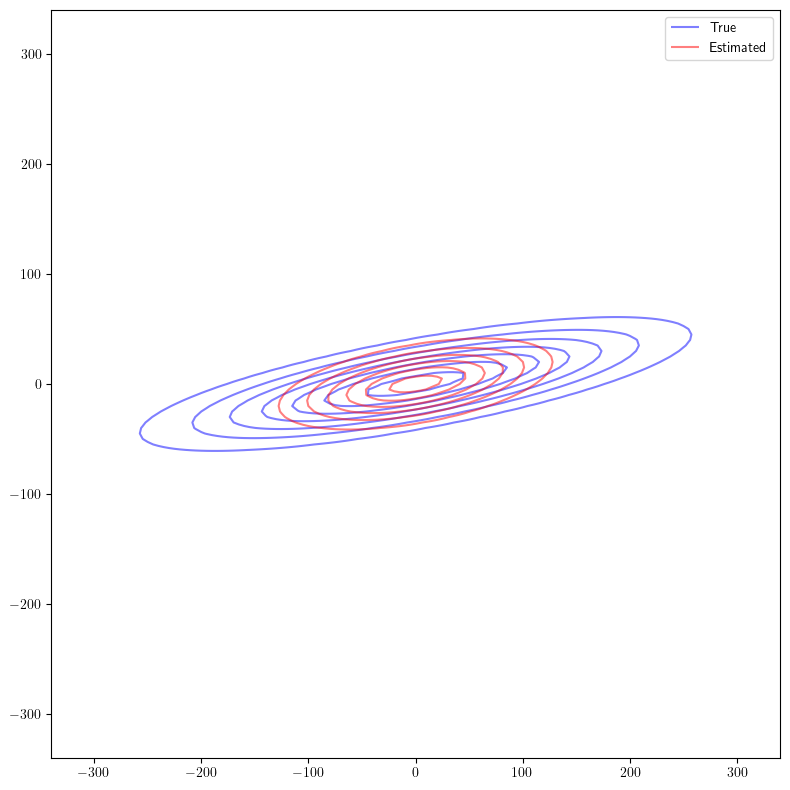}
            \label{fig:d3}
             \caption{$JD = 1.0389$}
        \end{subfigure}
        \hspace{2em}
        \begin{subfigure}[b]{0.3\columnwidth}
        \centering
            \includegraphics[width=\textwidth]{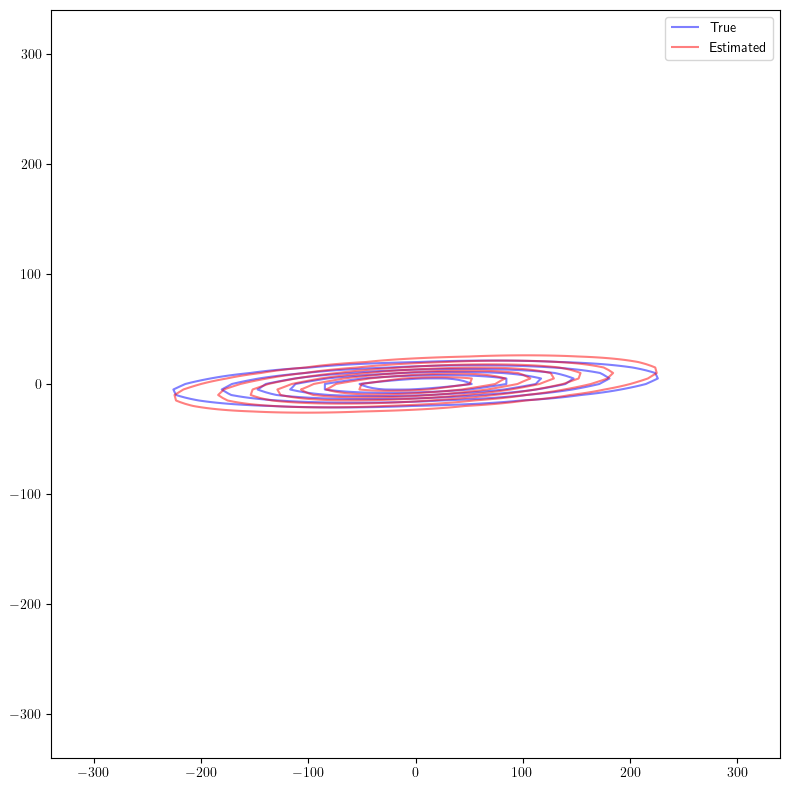}
            \label{fig:d4}
            \caption{$JD = 0.0846$}
        \end{subfigure}
 
        \caption{Jeffreys Divergence Scale Examples}
        \label{fig:distributionComparison}
    \end{figure}
    
\section{Experimental Results} \label{sec:exp-results}
    Two additional sets of experiments were conducted to compare the \mcse{} estimator to JEEDS, the previous best-performing execution skill estimation method. 
    One set focused on agents with arbitrary, randomly generated execution skill covariance matrices, but whose skill didn't change during the experiment. 
    The other set used agents whose execution skill varied over the course of the observations, to explore how effectively \mcse{} can track the skill of dynamic agents. 
    Details, results, and findings for each are presented next.

    \begin{figure}[ht]
            \centering
            \begin{subfigure}[b]{0.4\textwidth}
                \includegraphics[width=\textwidth]{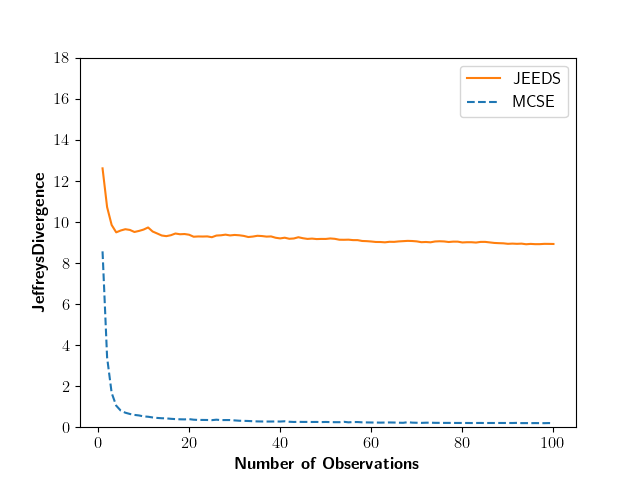}
                \label{fig:2dxe-rational}
                \caption{ Rational Agent }
            \end{subfigure}
            \begin{subfigure}[b]{0.4\textwidth}
                \includegraphics[width=\textwidth]{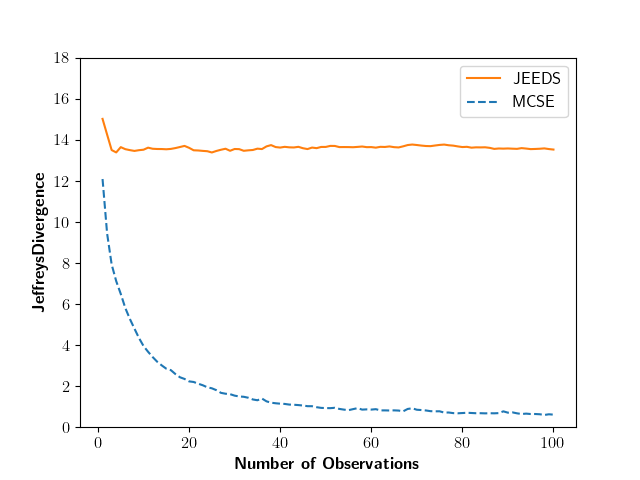}
                \label{fig:2dxe-softmax}
                \caption{ Softmax Agent }
            \end{subfigure}
            \begin{subfigure}[b]{0.4\textwidth}
                \includegraphics[width=\textwidth]{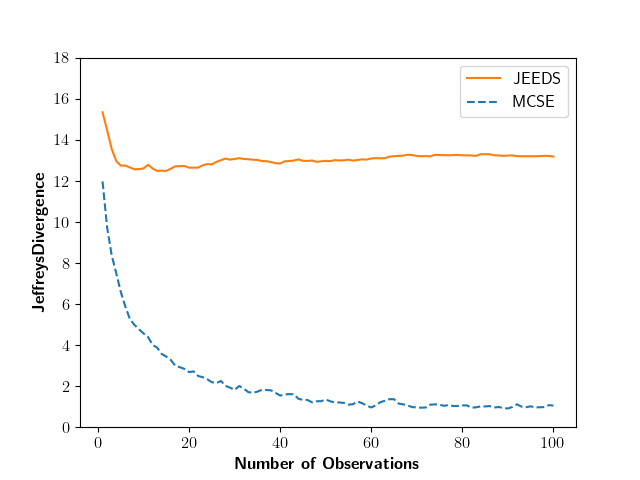}
                \label{fig:2dxe-flip}
                \caption{ Flip Agent }
            \end{subfigure}
            \begin{subfigure}[b]{0.4\textwidth}
                \includegraphics[width=\textwidth]{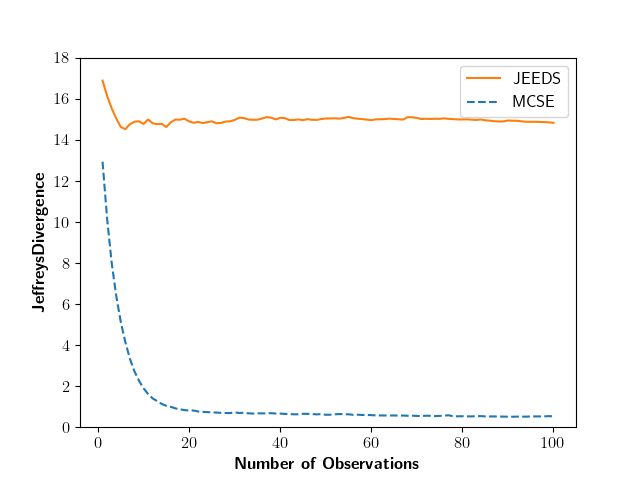}
                \label{fig:2dxe-deceptive}
                \caption{ Deceptive Agent }
            \end{subfigure}
        \caption{JD of Execution Skill Estimates in 2D-Darts}
        \label{fig:2d-x-jd}
    \end{figure}

    \subsection{Estimating Execution Skill in Higher Dimensions} \label{exps:multi-dimensions}
    We first investigate the performance of \mcse{} at estimating execution skill in higher dimensions.
    To do this, the following process was repeated.
    First, a random set of execution skill parameters ($\sigma_x, \sigma_y, \rho$) were generated, each from their respective ranges. 
    This execution skill level was used by all agents, who additionally each had a random decision-making skill level assigned, again from each agent's respective range of rationality parameters.
    These agents faced 100 randomly generated 2D-Darts states, and each estimator observed the sequence of states and executed actions, producing their execution skill estimate after each observation.
    At least 4000 agents of each type were included in these experiments. 

    Figure \ref{fig:2d-x-jd} shows the average JD for each method over observations, where the average is taken across all experiments involving that agent. 
    It is immediately clear that the performance of the \mcse{} estimators is significantly more accurate than JEEDS when used on agents with arbitrary execution noise distributions. 
    The average Jeffreys divergence for the \mcse{} estimators converges to values between 1 and 2 after the 100 observations for all the different agents, while the JEEDS average JD only nears 9 for the Rational agent, but is between 12 and 15 for the other agent types.
    To give some sense of the scale of these JD values, a visualization is given in Figure \ref{fig:distributionComparison}, showing
    pairs of distributions and their corresponding JD.

    \begin{figure}[H]
        \centering
        \begin{subfigure}[b]{0.4\textwidth}
            \includegraphics[width=\textwidth]{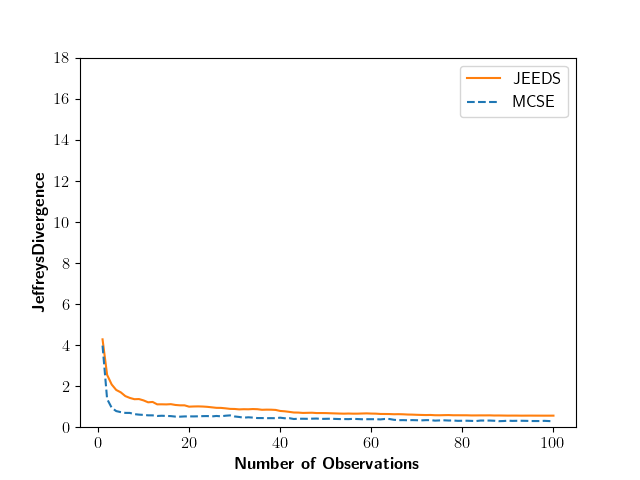}
            \label{fig:2dxe-rational-symmetric}
            \caption{ Symmetric Rational Agents }
        \end{subfigure}
        \begin{subfigure}[b]{0.4\textwidth}
            \includegraphics[width=\textwidth]{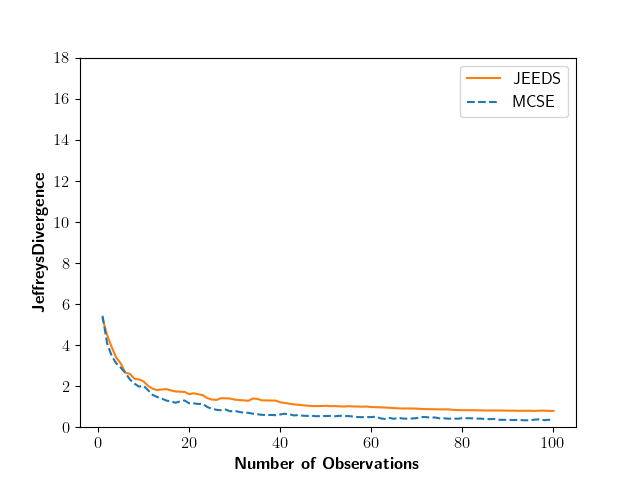}
            \label{fig:2dxe-sofmax-symmetric}
            \caption{ Symmetric Softmax Agent }
        \end{subfigure}
        \begin{subfigure}[b]{0.4\textwidth}
            \includegraphics[width=\textwidth]{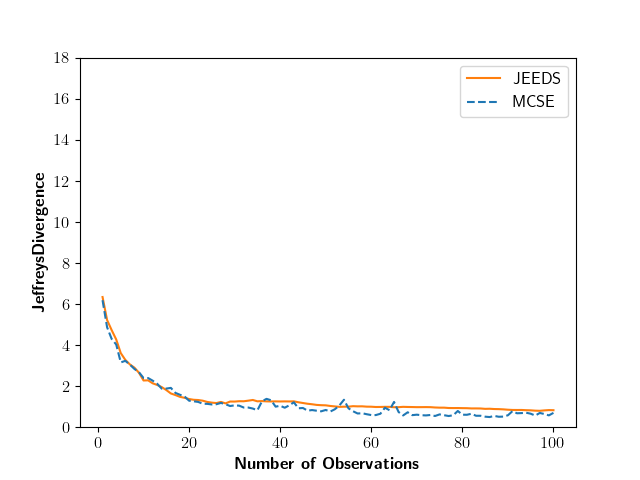}
            \label{fig:2dxe-flip-symmetric}
            \caption{ Symmetric Flip Agents }
        \end{subfigure}
        \begin{subfigure}[b]{0.4\textwidth}
            \includegraphics[width=\textwidth]{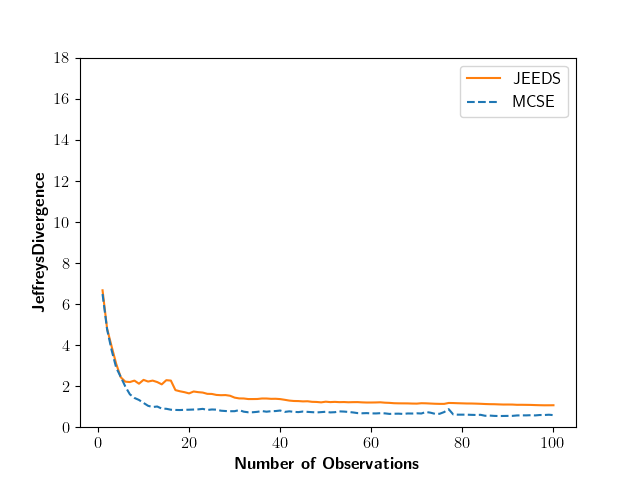}
            \label{fig:2dxe-tricker-symmetric}
            \caption{ Symmetric Deceptive Agent }
        \end{subfigure}
        \hfill
        \begin{subfigure}[b]{0.4\textwidth}
            \includegraphics[width=\textwidth]{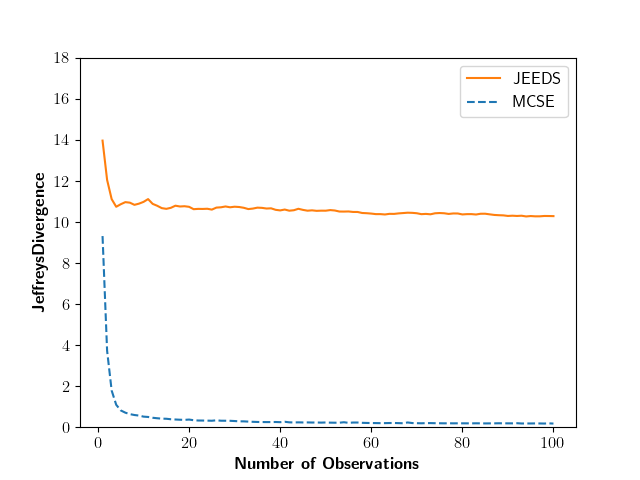}
            \label{fig:2dxe-rational-asymmetric}
            \caption{ Asymmetric Rational Agent }
        \end{subfigure}
        \begin{subfigure}[b]{0.4\textwidth}
            \includegraphics[width=\textwidth]{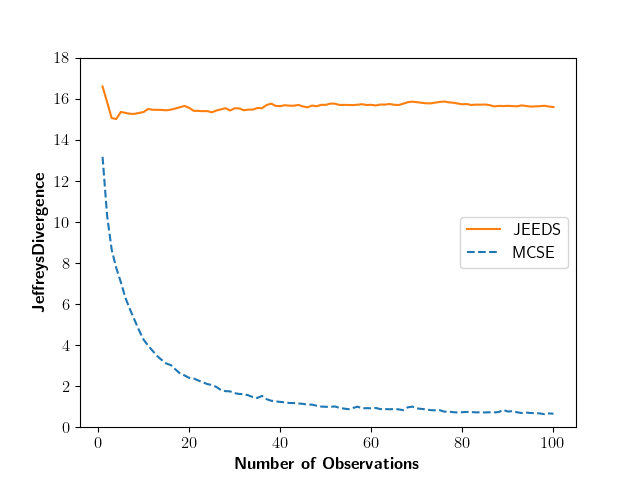}
            \label{fig:2dxe-sofmax-asymmetric}
            \caption{ Asymmetric Softmax Agent }
        \end{subfigure}
        \begin{subfigure}[b]{0.4\textwidth}
            \includegraphics[width=\textwidth]{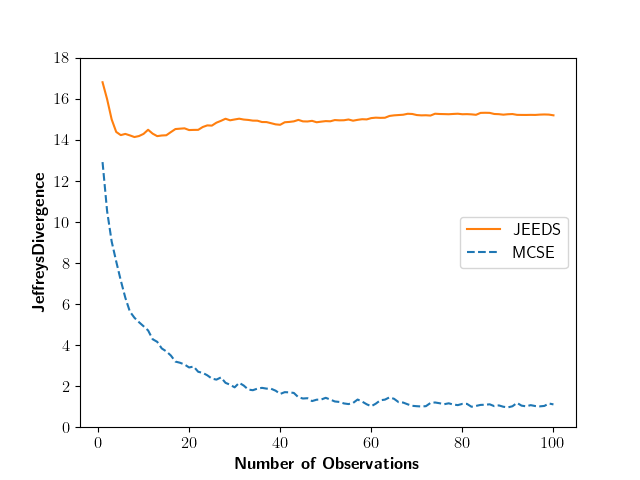}
            \label{fig:2dxe-flip-asymmetric}
            \caption{ Asymmetric Flip Agent }
        \end{subfigure}
        \begin{subfigure}[b]{0.4\textwidth}
            \includegraphics[width=\textwidth]{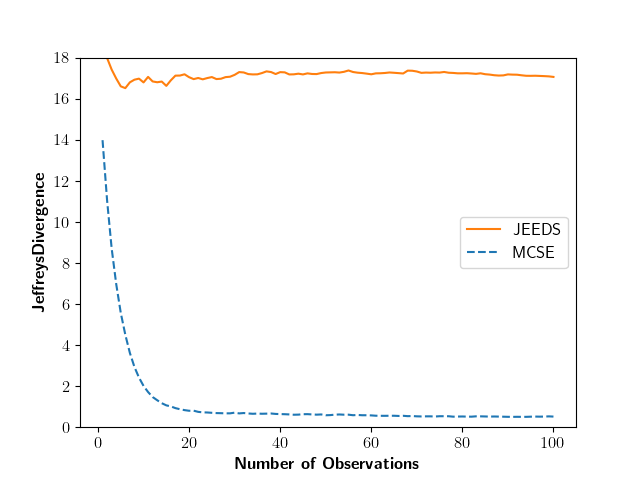}
            \label{fig:2dxe-tricker-asymmetric}
            \caption{ Asymmetric Deceptive Agent }
        \end{subfigure}
            
        \caption{JD of Execution Skill Estimates in 2D-Darts}
        \label{fig:2d-x-jd-symmetric-vs-asymmetric}
    \end{figure}
 
    To shed more light on where the improvement is coming from, Figure \ref{fig:2d-x-jd-symmetric-vs-asymmetric} separates the agents into two groups, those with more symmetric execution noise distributions and the others.  
    There were 14\% of the agents that we categorized as symmetric, with $|\sigma_x-\sigma_y| < 50$ and $|\rho| < 0.2$. 
    It would make sense for JEEDS to do better on the symmetric agents, as those agents match the assumption of symmetry made by the JEEDS method. 
    The symmetric results, shown on the top row of Figure \ref{fig:2d-x-jd-symmetric-vs-asymmetric}, show that \mcse{} is competitive with and often barely outperforms JEEDS on these agents.  
    The average JD for the both estimators is still good, converging to values below 2. 
    The asymmetric results show that these are the cases where JEEDS really struggles and \mcse{} does very well, with the plots closely resembling the overall plots shown in Figure \ref{fig:2d-x-jd}.

    From these experiments, we conclude that the \mcse{} is much more effective at estimating execution skill in higher dimensions than the previous methods. 
    It can do this consistently, even with agents who are not perfectly rational in their action selection mechanism.

    \subsection{Estimating Execution Skill Over Time} \label{exps:dynamic-xskill}
    We next explore the ability of \mcse{} to estimate execution skill as it changes over time. 
    To do this, we made variations of the \emph{Rational} and the \emph{Softmax} agents, with two different ways that skill can change over time. 
    Each require initial and final skill levels to be specified, with one agent type changing skill abruptly at a single observation, and the other changing skill gradually across all observations.    
    Abrupt changes are made at a random point in the middle third of the observations (between 33 and 66 with 100 observations). 
    The execution skill level for a given dynamic agent was created as follows:
    First, $[8.0, 15.0]$ mm and $[130.0, 145.0]$ mm were defined as representative ranges for agents with more accurate and less accurate execution skills, respectively.
    Then, two samples were drawn from each range (one for $\sigma_x$ and one for $\sigma_y$, along with a random $\rho$ value) to create the initial and final skill levels for the agent.
    The orders in which these were assigned varied as the agent's skill level could progress from more accurate to less accurate or vice-versa.

    \begin{figure}[ht]
        \centering
        \begin{subfigure}[b]{0.4\textwidth}
            \includegraphics[width=\textwidth]{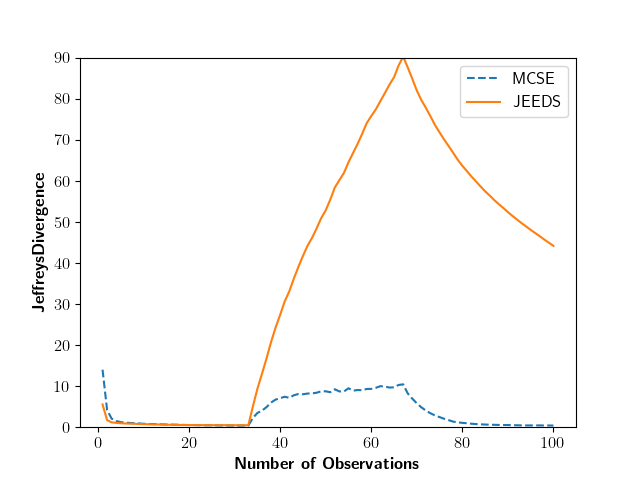}
            \label{fig:2dxe-rational-abrupt}
            \caption{ Abrupt Rational Agent }
        \end{subfigure}%
        \hspace{2em}
        \begin{subfigure}[b]{0.4\textwidth}
            \includegraphics[width=\textwidth]{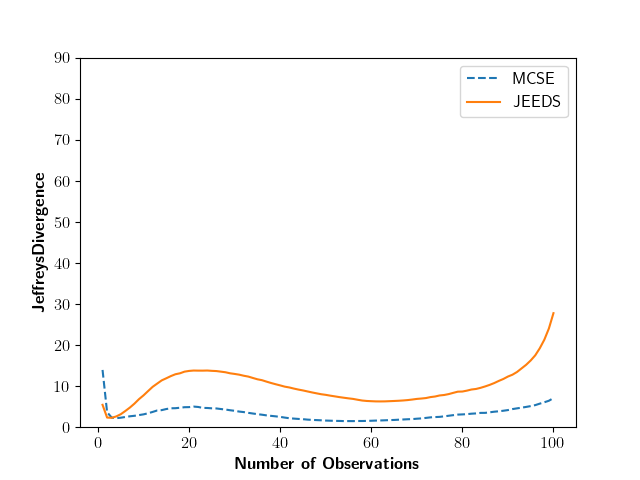}
            \label{fig:2dxe-rational-gradual}
            \caption{ Gradual Rational Agent }
        \end{subfigure}%
        
        \begin{subfigure}[b]{0.4\textwidth}
            \includegraphics[width=\textwidth]{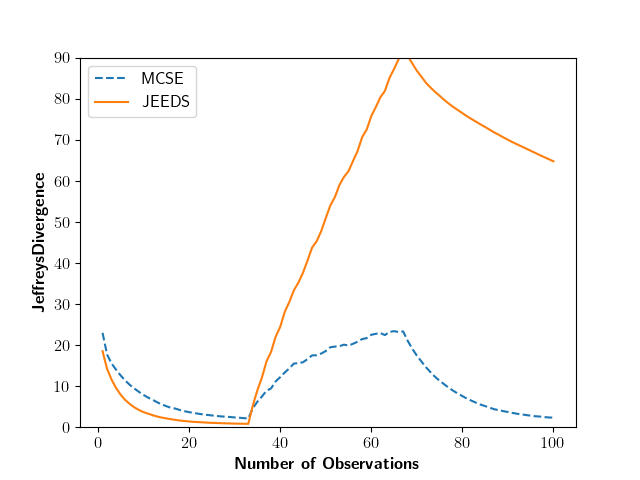}
            \label{fig:2dxe-softmax-abrupt}
            \caption{ Abrupt Softmax Agent }
        \end{subfigure}%
        \hspace{2em}
        \begin{subfigure}[b]{0.4\textwidth}
            \includegraphics[width=\textwidth]{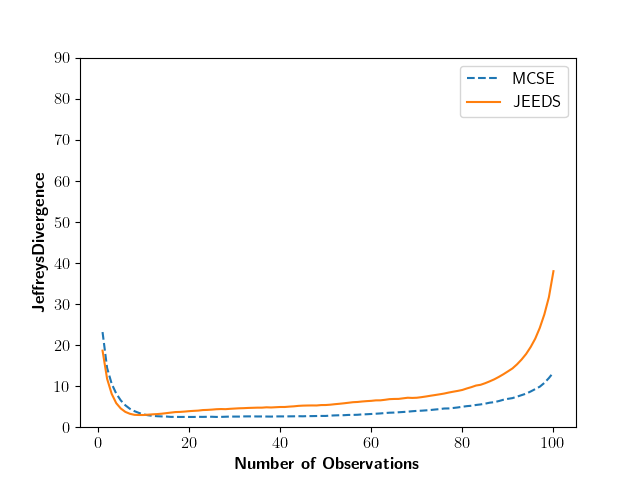}
            \label{fig:2dxe-softmax-gradual}
            \caption{ Gradual Softmax Agent }
        \end{subfigure}
    \caption{JD in 2D-Darts for Dynamic Agents}
    \label{fig:2d-x-jd-dynamic-agents}
    \end{figure}

    \mcse{} again exhibits smaller errors than JEEDS at this task.
    The abrupt execution skill changes cause massive spikes in JD for JEEDS, from which it never completely recovers. 
    While \mcse{} also has more error around the changes, it is of lower magnitude and the estimate quickly recovers afterwards.
    The gradual agents present more of a challenge, as the \mcse{} estimators reach average JD error levels of over 5. 
    This is still less than the error of the JEEDS method.
    Perhaps more troubling is the fact that for the gradual agents, as the end of the observations nears, the error is ever-increasing. 
    This is perhaps understandable, as the agent is always changing its execution skill level, and this is perhaps a greater challenge for both methods than we anticipated. 
    Overall, \mcse{} shows much improvement over JEEDS at estimating dynamic skill, but there appears to still be room for improvement in the gradual changing skill case.
    
    Experiments of \mcse{} with two different $w\%$ values were also performed to investigate whether the amount of noise affected the performance of the estimators on this task. 
    Results, found in Appendix \ref{apx:noise} show that the noise did not significantly affect their performance.

\section{Application: Skill in Baseball} \label{sec:baseball}
We now demonstrate the applicability of \mcse{} to real-world Major League Baseball (MLB) data. 
In baseball, a pitcher attempts to throw a baseball so that a batter cannot successfully hit it.\footnote{For complete rules see \url{https://www.rulesofsport.com/sports/baseball.html}}
 The execution skill of the pitcher is an important factor in the pitcher's success. 
Public data exists that gives the location of each pitch \cite{pybaseballWebsite}. 
We use the same techniques as \cite{melville2023baseball} to generate a reward function over the space of pitch locations for each pitch. 
The walks per inning (BB/IP) statistic is generally considered to be a measure of a pitcher's accuracy. 
Using data from 2021, we selected three top-ranked pitchers by BB/IP and three bottom-ranked pitchers, each with at least 100 fastball (FF) pitches that we could feed into the \mcse{} and JEEDS estimators. 
The generalized variance (GV) (determinant of the estimated covariance matrix) for each pitcher on the FF pitch type is shown in Table \ref{tab:baseball}, with the top-ranked pitchers shown above the bottom-ranked.
This number gives a sense of the spread of a bivariate normal distribution, where higher numbers indicate less accuracy.

\begin{table}[]
    \caption{Execution Skill Estimates for MLB Pitchers}
    \label{tab:baseball}
    \centering
    \begin{tabular}{cc|cc|cc}
    \toprule
    \multicolumn{2}{c|}{} & \multicolumn{2}{c|}{GV} & \multicolumn{2}{c}{\% in Strike Zone} \\
    \hline 
    Pitcher & BP/IP & JEEDS & \mcse{} & JEEDS & \mcse{} \\ 
    \hline
    Chris Martin & 0.091 & 0.162 & 0.019 & 63.25 & 94.51 \\
    Jacob deGrom & 0.124 & 0.278 & 0.011 & 53.84 & 97.69 \\
    Corey Kluber & 0.128 & 0.585 & 0.030 & 41.69 & 89.96 \\
    \hline
    Tanner Scott & 0.741 & 0.597 & 0.024 & 41.25 & 90.43 \\
    Jake Diekman & 0.736 & 0.433 & 0.153 & 46.46 & 63.40 \\
    Tucker Davidson & 0.673 & 0.424 & 0.017 & 46.75 & 95.10 \\
    \bottomrule
    \end{tabular}
\end{table}

\begin{figure}
    \centering
    \includegraphics[width=0.4\columnwidth]{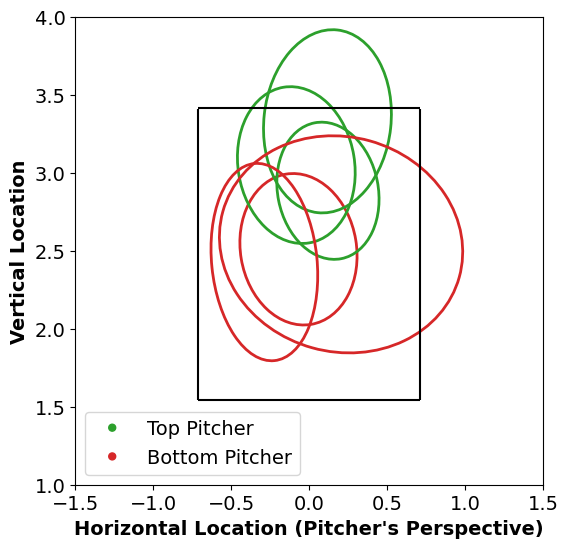}
    \caption{\mcse{} estimated MLB Execution Error Distributions}
    \label{fig:baseball}
\end{figure}

The methods were generally able to distinguish between the pitchers in the two groups (the average GV for the top pitchers is lower than for the bottom pitchers for each estimator), and the estimates vary significantly between estimators.  
We are unable to state which estimator is more accurate, as the ground truth error distributions for the pitchers are unknown.
Table \ref{tab:baseball}, also shows, in the rightmost column, our estimated probability of a pitch being in the strike zone for each pitcher, given they were aimed in exact middle.  
This was estimated using one million samples from each estimated distribution.
Figure \ref{fig:baseball} shows the 50\% confidence ellipses for the estimates produced by the \mcse{} estimator, giving a sense as to the shape of the estimated error distributions.
Each ellipse is centered on the average fastball pitch location for that pitcher. 
Most of the estimates have more variance in the y dimension than the x dimension, showing the value of estimating execution skill in a higher dimension.

\section{Conclusions} \label{sec:conclusion}
In this paper, the Monte Carlo Skill Estimation (\mcse{}) was introduced, which utilizes a particle filter to perform inference in a dynamic Bayesian network with random variables corresponding to the execution and decision-making skill of an observed agent.
\mcse{} makes two contributions: 1) it can model and estimate execution skill in multiple dimensions, and 2) it can better handle agents whose execution skill varies over time.
These properties make it a natural method to use on real-world data, as was demonstrated on MLB data. 
In the future, we plan to explore additional enhancements and extensions of the basic particle filter approach that was used in this paper, given the success of this initial exploration.
We also plan to apply the \mcse{} method to additional real-world data sources. 

\bibliography{refs}
\bibliographystyle{abbrv}

\begin{appendices}

\section{Preliminary Experiments} \label{apx:prelim}
This section presents details for the two different rounds of experiments conducted to systematically explore a range of parameter values with the goal of identifying optimal parameters to be used for subsequent experiments presented in the paper.

A total of 9 different stationary \textit{Rational} agents were evaluated in each round.
These agents presented execution skill levels of $(10,10) mm$, $(10,100) mm$, and $(100,100) mm$.
Three $\rho$ parameters $(-0.75, 0.0, 0.75)$ were used with each execution skill level.
This allowed for the set of agents to have samples from symmetric agents and asymmetric ones.
All of the agents were seen for 100 observations in the 2D-Darts domain.

    \subsection{Round 1: Parameter Sweep}
    The goal of this round of experiments was to determine an ideal value for the motion model noise $w_{\%}$, 
    select the most effective resampling strategy (whether to employ constant resampling or to use \neff{}, and identify an ideal percentage of particles to resample ($r$).

    A total of 18 different estimators were evaluated.
    Table \ref{table:paramSweep} presents the different combinations used for the parameters of the estimators along with their final JD value.
    Note that they are sorted from smallest to highest JD.
    The selected estimator (in bold) was $w_{\%} = 0.005$, $r = 90$, with \neff{}.

    \begin{table}[ht]
        \centering
        \begin{tabular}{|ccc|c|}
        \toprule
        \multicolumn{3}{|c|}{Estimator} & \multicolumn{1}{c|}{JD} \\
        $w_{\%}$ & $r$ & \neff{}? &  \\
        \hline
        \textbf{0.005} & \textbf{90} &  \textbf{yes} & \textbf{0.2052}\\
        0.005 & 95 & yes & 0.2064 \\
        0.005 & 75 & yes & 0.2339 \\
        0.002 & 75 & yes & 0.2366 \\
        0.005 & 95 & no & 0.2451 \\
        0.002 & 90 & yes & 0.2465 \\
        0.005 & 90 & no & 0.2579 \\
        0.002 & 90 & no & 0.2706 \\
        0.002 & 95 & yes & 0.2787 \\
        0.020 & 95 & yes & 0.3002 \\
        0.020 & 90 & yes & 0.3167 \\
        0.020 & 95 & no & 0.3471 \\
        0.002 & 95 & no & 0.3526 \\
        0.020 & 90 & no & 0.4018 \\
        0.020 & 75 & yes & 0.4314 \\
        0.005 & 75 & no & 0.4436 \\
        0.002 & 75 & no & 0.4656 \\
        0.020 & 75 & no & 0.6821 \\
        \bottomrule
        \end{tabular}
        \caption{Evaluation of the Estimators}
        \label{table:paramSweep}
    \end{table}

    \subsection{Round 2: Setting the Number of Particles}
    The goal of this round of experiments was to determine an ideal value for the number of particles to use ($M$) within a given estimator.
    
\begin{figure}[H]
        \centering
        \includegraphics[width=0.6\columnwidth]{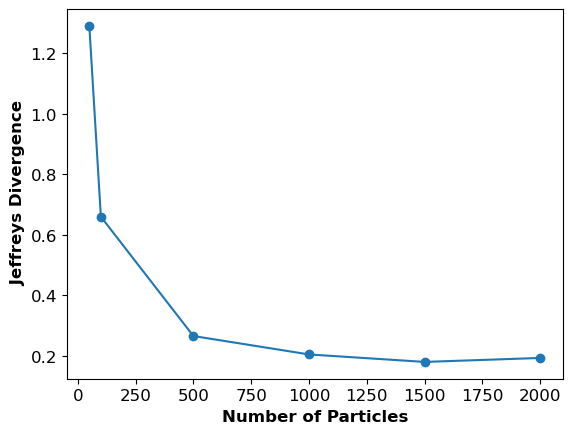}
        \caption{}
        \label{fig:resultsSettingM}
    \end{figure}

    Experiments were conducted with the best-performing estimator from the previous round ($w_{\%} = 0.005$, $r = 90$, with \neff{}) while varying $M$ in $[$50, 100, 500, 1000, 1500, 2000$]$.
    Table \ref{table:statsRound2} presents the total number of experiments obtained for each selected agent across different $M$'s.
    Figure \ref{fig:resultsSettingM} shows how the final JD changes as $M$ increases.
    $M=1000$ was selected for subsequent experiments as its performance is comparable to that of $M=1500$ and $M=2000$, while also reducing the running time of the experiments.
    
    \begin{table}[ht]
        \centering
        \begin{tabular}{|c|cccccc|}
            \toprule
            & \multicolumn{6}{c|}{Number of Particles} \\
            Rational Agent & 50 & 100 & 500 & 1000 & 1500 & 2000 \\
            \hline
            (10,10,-0.75) & 1000 & 974 & 1147 & 1151 & 1095 & 941 \\
            (10,10,0.0) & 1000 & 977 & 1159 & 1183 & 1128 & 986 \\
            (10,10,0.75) & 1000 & 975 & 1153 & 1168 & 1106 & 978 \\
            (10,100,-0.75) & 953 & 945 & 936 & 1111 & 975 & 1043 \\
            (10,100,0.0) & 955 & 948 & 947 & 1142 & 1007 & 1090 \\
            (10,100,0.75) & 955 & 947 & 942 & 1123 & 992 & 1079 \\
            (100,100,-0.75) & 946 & 937 & 916 & 952 & 936 & 1220 \\
            (100,100,0.0) & 947 & 940 & 928 & 974 & 961 & 1266 \\
            (100,100,0.75) & 946 & 939 & 926 & 956 & 945 & 1249 \\
            \hline
            Total & 8702 & 8582 & 9054 & 9760 & 9145 & 9852 \\
            \bottomrule
        \end{tabular}
        \caption{Experiments Conducted - Round 2}
         \label{table:statsRound2}
    \end{table}

\onecolumn

\section{Experiments Performed: Different Agent Types}

    \begin{table}[ht]
        \centering
        \begin{tabular}{|c|c|}
            \toprule
            Agents & Total \\
            \hline
            Rational & 4255 \\
            Softmax &  4092 \\
            Flip & 4196 \\
            Deceptive & 4129 \\
            \hline
            Abrupt Rational & 5838 \\
            Gradual Rational &  5942 \\
            Abrupt Softmax & 5633 \\
            Gradual Softmax & 5948 \\
            \bottomrule
        \end{tabular}
        \caption{Experiments Conducted}
         \label{table:statsRandExps}
    \end{table}

\section{Experiments Performed: Dynamic Agents - Varying $w\%$} \label{apx:noise}

    \begin{figure}[ht]
        \centering
        \begin{subfigure}[b]{0.4\columnwidth}
            \includegraphics[width=\textwidth]{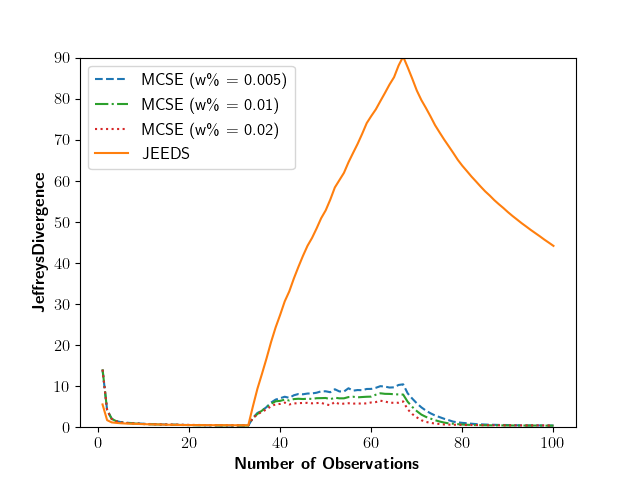}
            \label{fig:2dxe-target-abrupt-allNoises}
            \caption{ Abrupt Rational Agent }
        \end{subfigure}
        \begin{subfigure}[b]{0.4\columnwidth}
            \includegraphics[width=\textwidth]{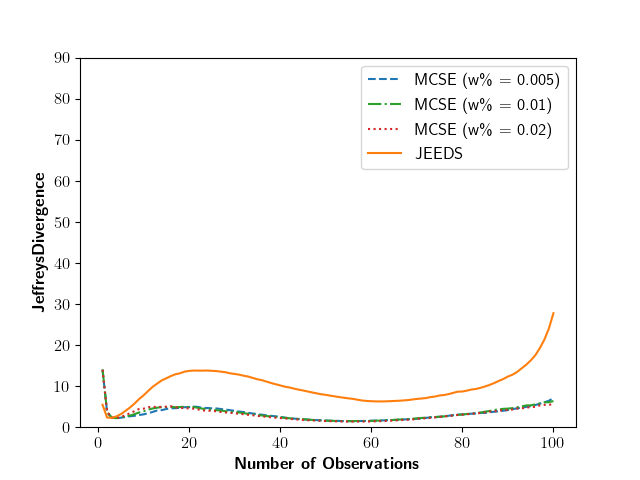}
            \label{fig:2dxe-target-gradual-allNoises}
            \caption{ Gradual Rational Agent }
        \end{subfigure}
        \begin{subfigure}[b]{0.4\columnwidth}
            \includegraphics[width=\textwidth]{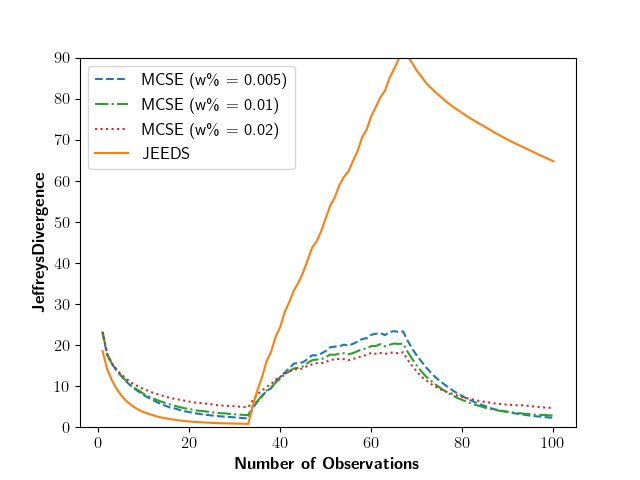}
            \label{fig:2dxe-softmax-abrupt-allNoises}
            \caption{ Abrupt Softmax Agent }
        \end{subfigure}
        \begin{subfigure}[b]{0.4\columnwidth}
            \includegraphics[width=\textwidth]{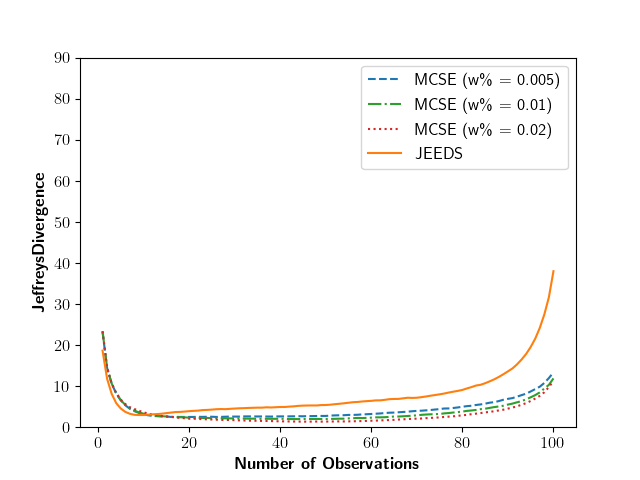}
            \label{fig:2dxe-softmax-gradual-allNoises}
            \caption{ Gradual Softmax Agent }
        \end{subfigure}
    \caption{JD in 2D-Darts for Dynamic Agents - Different $w\%$'s}
    \label{fig:2d-x-jd-dynamic-agents-allNoises}
    \end{figure}

\clearpage 
\section{Sample Distributions for Different Agent Types}

    \begin{figure}[ht]
        \centering
        \begin{subfigure}[b]{0.4\columnwidth}
            \includegraphics[width=\textwidth]{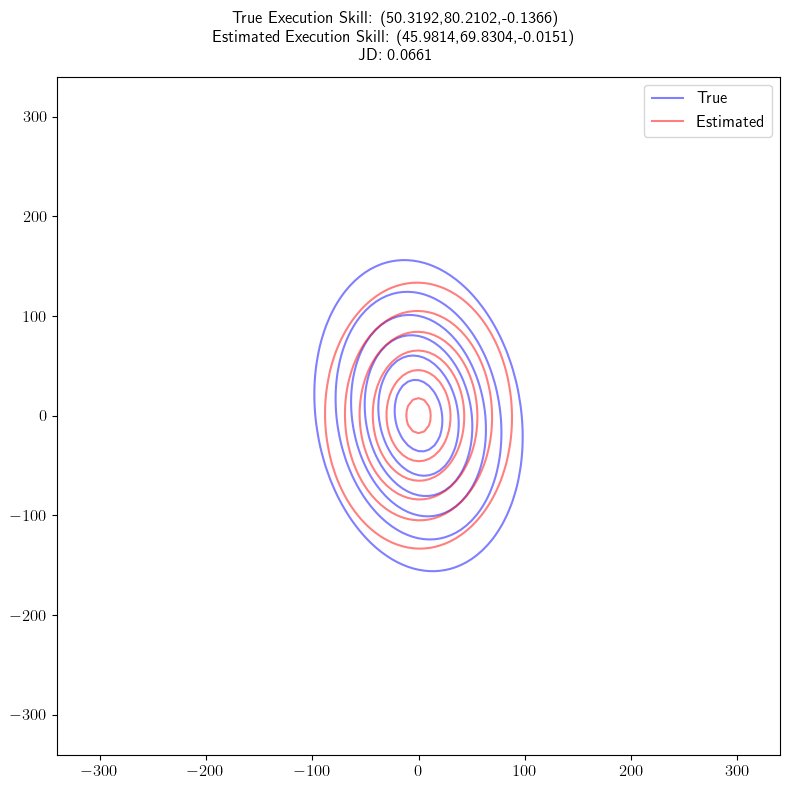}
        \end{subfigure}
        \begin{subfigure}[b]{0.4\columnwidth}
        \includegraphics[width=\textwidth]{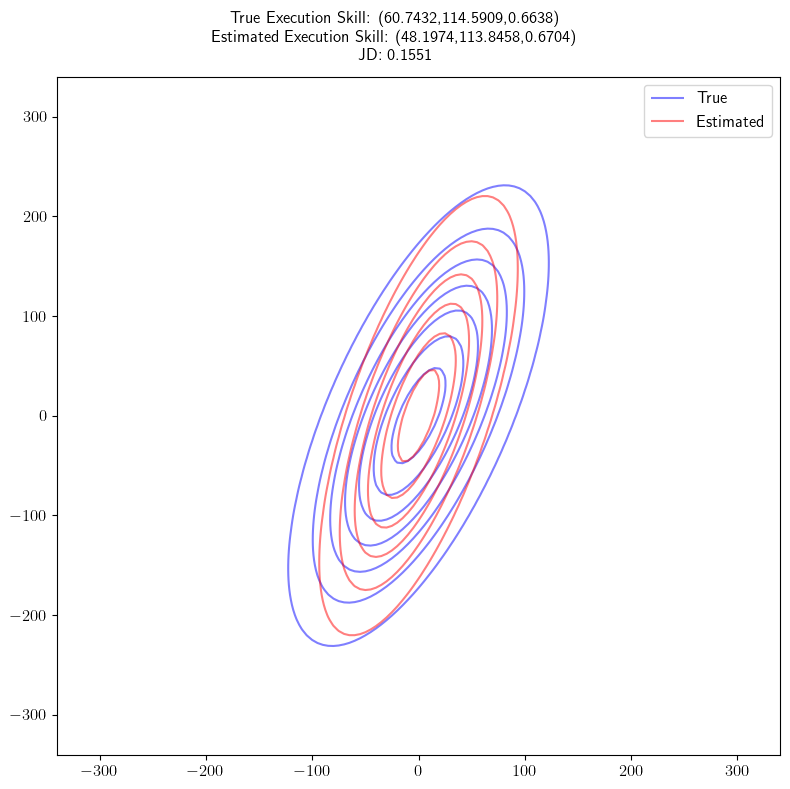}
        \end{subfigure}
        \caption{Sample Rational Agents}
        \label{fig:distRational}
    \end{figure}

    \begin{figure}[ht]
        \centering
        \begin{subfigure}[b]{0.4\columnwidth}
            \includegraphics[width=\textwidth]{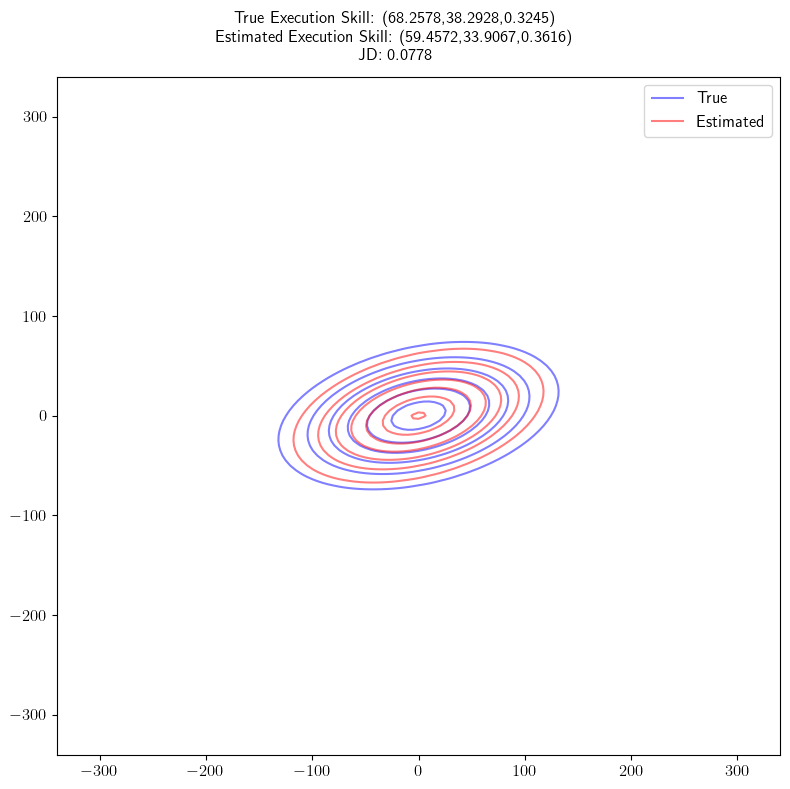}
        \end{subfigure}
        \begin{subfigure}[b]{0.4\columnwidth}
            \includegraphics[width=\textwidth]{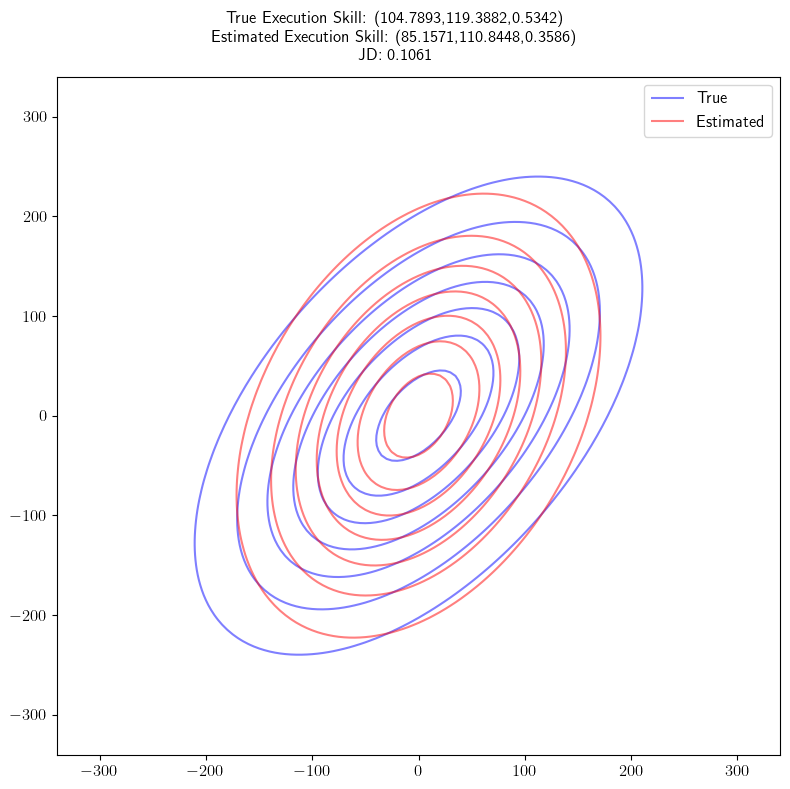}
        \end{subfigure}
        \caption{Sample Softmax Agents}
        \label{fig:distBounded}
    \end{figure}

    \begin{figure}[ht]
        \centering
        \begin{subfigure}[b]{0.4\columnwidth}
            \includegraphics[width=\textwidth]{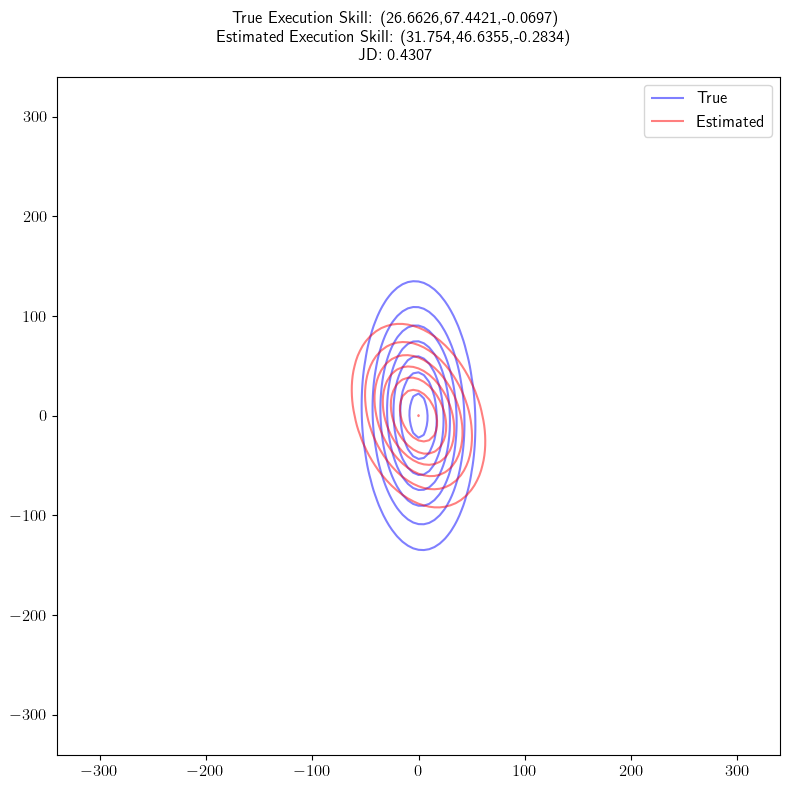}
        \end{subfigure}
        \begin{subfigure}[b]{0.4\columnwidth}
            \includegraphics[width=\textwidth]{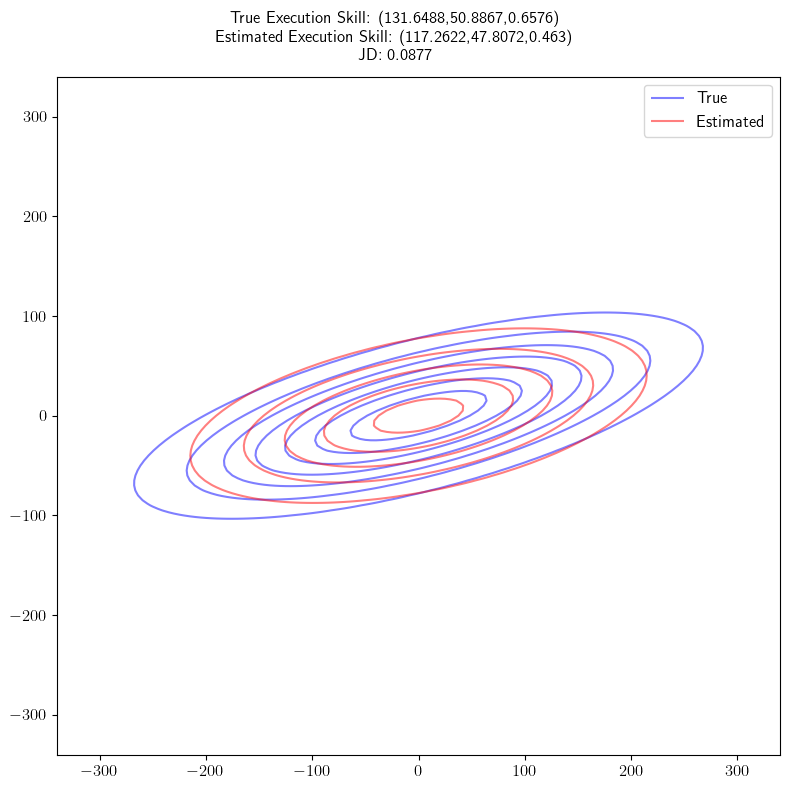}
        \end{subfigure}
        \caption{Sample Flip Agents}
        \label{fig:distFlip}
    \end{figure}

    \begin{figure}[ht]
        \centering
        \begin{subfigure}[b]{0.4\columnwidth}
            \includegraphics[width=\textwidth]{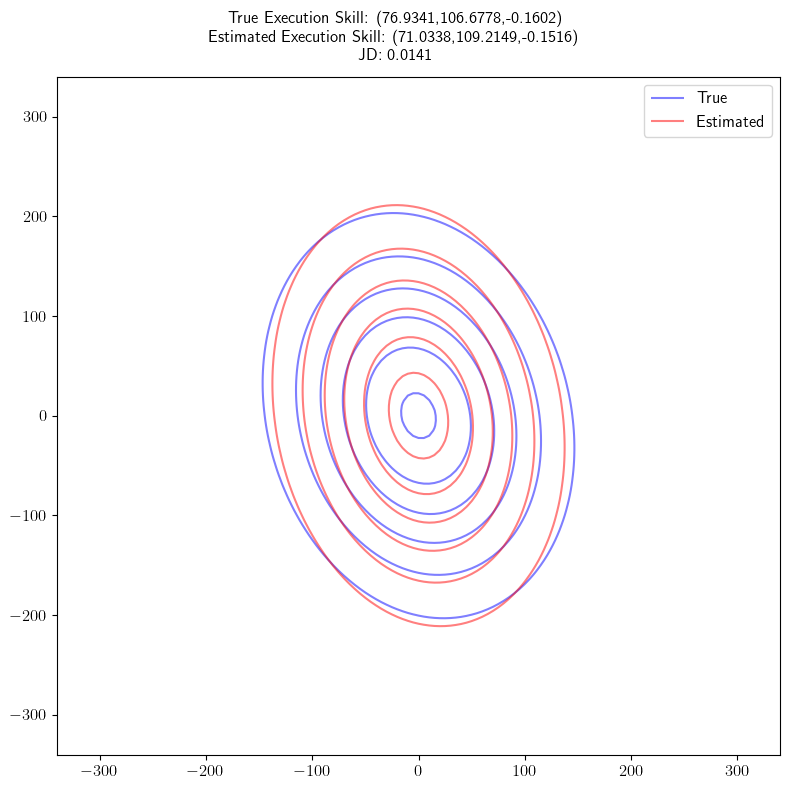}
        \end{subfigure}
        \begin{subfigure}[b]{0.4\columnwidth}
            \includegraphics[width=\textwidth]{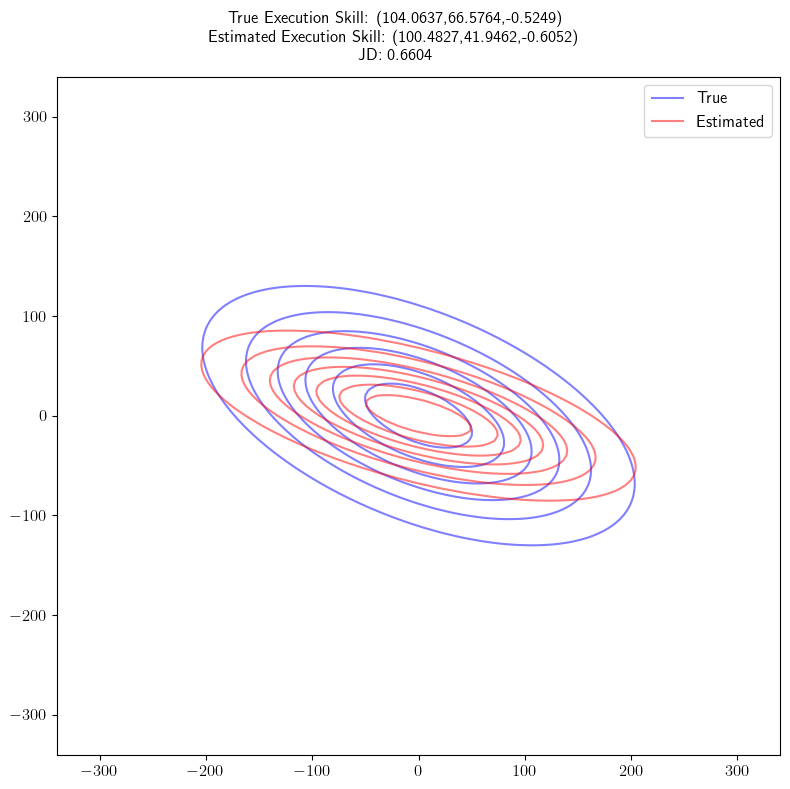}
        \end{subfigure}
        \caption{Sample Deceptive Agents}
        \label{fig:distTricker}
    \end{figure}

\end{appendices}

\end{document}